\documentclass[pre, twocolumn,amsmath]{revtex4-1}

\usepackage{graphicx}
\usepackage{amssymb,amsfonts,amsmath}

\def \>{\rangle} 
\def \<{\langle} 
 
\def\be{\begin{equation}} 
\def\ee{\end{equation}} 
\def\longrightharpoonup{\relbar\joinrel\rightharpoonup}
\def\longleftharpoondown{\leftharpoondown\joinrel\relbar}

\def\longrightleftharpoons{
  \mathop{
    \vcenter{
      \hbox{
      \ooalign{
        \raise1pt\hbox{$\longrightharpoonup\joinrel$}\crcr
	  \lower1pt\hbox{$\longleftharpoondown\joinrel$}
	  }
      }
    }
  }
}

\newcommand \bea {\begin{eqnarray}} 
\newcommand \eea {\end{eqnarray}}

\begin{document}

\title{An exact mapping between the Variational Renormalization Group and Deep Learning}

\author{Pankaj Mehta}
\affiliation{Dept. of Physics, Boston University, Boston, MA}
\author{David J. Schwab}
\affiliation{Dept. of Physics, Northwestern University, Evanston, IL}

\begin{abstract}
{Deep learning is a broad set of techniques that uses multiple layers of representation to automatically learn relevant features directly from structured data. Recently, such techniques have yielded record-breaking results on a diverse set of difficult machine learning tasks in computer vision, speech recognition, and natural language processing.   Despite the enormous success of deep learning, relatively little is understood theoretically about why these techniques are so successful at feature learning and compression. Here, we show that deep learning is intimately related to one of the most important and successful techniques in theoretical physics, the renormalization group (RG). RG is an iterative coarse-graining scheme that allows for the extraction of relevant features (i.e. operators) as a physical system is examined at different length scales. We construct an exact mapping from the variational renormalization group, first introduced by Kadanoff, and deep learning architectures based on Restricted Boltzmann Machines (RBMs). We illustrate these ideas using the nearest-neighbor Ising Model in one and two-dimensions.  Our results suggests that deep learning algorithms may be employing a generalized RG-like scheme to learn relevant features from data.}
\end{abstract}

\maketitle

A central goal of modern machine learning research is to learn and extract important features directly from data. Among the most promising and successful techniques for accomplishing this goal are those associated with the emerging sub-discipline of deep learning. Deep learning uses multiple layers of representation to learn descriptive features directly from training data \cite{bengio2009learning, 38115} and has been successfully utilized, often achieving record breaking results, in difficult machine learning tasks including object labeling \cite{imagenet}, speech recognition \cite{speechrecognition}, and natural language processing \cite{deepnlp}.

In this work, we will focus on a set of deep learning algorithms known as deep neural networks (DNNs) \cite{hinton2006reducing}. DNNs are biologically-inspired graphical statistical models that consist of multiple layers of ``neurons'', with units in one layer receiving inputs from units in the layer below them. Despite their enormous success, it is still unclear what advantages these deep, multi-layer architectures possess over shallower architectures with a similar number of parameters. In particular, it is still not well understood theoretically why DNNs are so successful at uncovering features in structured data. (But see \cite{bengiolecun,le2010,le2008representational}.)

One possible explanation for the success of DNN architectures is that they can be viewed as an iterative coarse-graining scheme, where each new high-level layer of the neural network learns increasingly abstract higher-level features from the data \cite{bengio2007greedy, bengio2009learning}.  The initial layers of the the DNN can be thought of as low-level feature detecters which are then fed into higher layers in the DNN which combine these low-level features into more abstract higher-level features, providing a useful, and at times reduced, representation of the data. By successively applying feature extraction, DNNs learn to deemphasize irrelevant features in the data while simultaneously learning relevant ones. (Note that in a supervised setting, such as classification, relevant and irrelevant are ultimately determined by the problem at hand. Here we are concerned solely with the unsupervised aspect of training DNNs, and the use of DNNs for compression \cite{hinton2006reducing}.) In what follows, we make this explanation precise.

This successive coarse-graining procedure is reminiscent of one of the most successful and important tools in theoretical physics, the renormalization group (RG) \cite{wilson1974renormalization, wilson1983renormalization}. RG is an iterative coarse-graining procedure designed to tackle difficult physics problems involving many length scales. The central goal of RG is to extract relevant features of a physical system for describing phenomena at large length scales by integrating out (i.e. marginalizing over) short distance degrees of freedom. In any RG sequence, fluctuations are integrated out starting at the microscopic scale and then moving iteratively on to fluctuations at larger scales. Under this procedure, certain features, called relevant operators, become increasingly important while other features, dubbed irrelevant operators, have a diminishing effect on the physical properties of the system at large scales.

In general, it is impossible to carry out the renormalization procedure exactly. Therefore, a number of approximate RG procedures have been developed in the theoretical physics community \cite{wilson1983renormalization, cardy1996scaling,kadanoff2000statics, goldenfeld1992lectures}. One such approximate method is a class of variational ``real-space'' renormalization schemes introduced by Kadanoff for performing RG on spin systems \cite{kadanoff2000statics, kadanoff1976variational, efrati2014real}.  Kadanoff's variational RG scheme introduces coarse-grained auxiliary, or ``hidden'', spins that are coupled to the  physical spin systems through some unknown coupling parameters. A parameter-dependent free energy is calculated for the coarse-grained spin system from the coupled system by integrating out the physical spins. The coupling parameters are chosen through a variational procedure that minimizes the difference between the free energies of the physical and hidden spin systems. This ensures that the coarse-grained system preserves the long-distance information present in the physical system. Carrying out this procedure results in an RG transformation that maps the physical spin system into a coarse-grained description in terms of hidden spins. The hidden spins then serve as the input for the next round of renormalization. 

The introduction of layers of hidden spins is also a central component of DNNs based on Restricted Boltzmann Machines (RBMs). In RBMs, hidden spins (often called units or neurons) are coupled to ``visible" spins describing the data of interest. (Here we restrict ourselves to binary data.) The coupling parameters between the visible and hidden layers are chosen using a variational procedure that minimizes the Kullback-Leibler divergence (i.e. relative entropy) between the ``true" probability distribution of the data and the variational distribution obtained by marginalizing over the hidden spins. Like in variational RG, RBMs can be used to map a state of the visible spins in a data sample into a description in terms of hidden spins. If the number of hidden units is less than the number of visible units, such a mapping can be thought of as a compression. (Note, however, that dimensional expansions are common \cite{bengio2013representation}.) In deep learning, individual RBMs are stacked on top of each other into a DNN  \cite{hinton2006reducing, hinton2006fast}, with the output of one RBM serving as the input to the next. Moreover, the variational procedure is often performed iteratively, layer by layer.

The preceding paragraphs suggest an intimate connection between RG and deep learning. Indeed, here we construct an {\it exact mapping} from the variational RG scheme of Kadanoff to DNNs based on RBMs \cite{hinton2006reducing, hinton2006fast}. Our mapping suggests that DNNs implement a generalized RG-like procedure to extract relevant features from structured data.

The paper is organized as follows. We begin by reviewing Kadanoff's variational renormalization scheme in the context of the Ising Model. We then introduce RBMs and deep learning architectures of stacked RBMs. We then show how to map the procedure of variational RG to unsupervised training of a DNN. We illustrate these ideas using the one- and two-dimensional nearest-neighbor Ising models. We end by discussing the implication of our mapping for physics and machine learning.  

\begin{figure*}
\center
\includegraphics{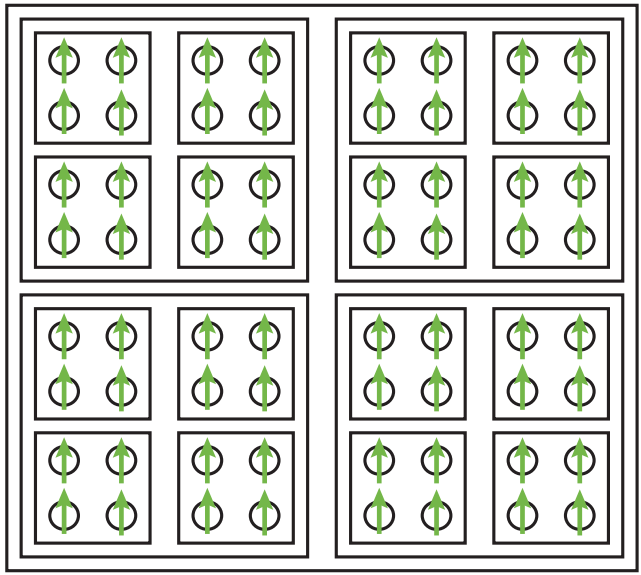}
\caption{{\bf Block spin renormalization.} In block spin renormalization \cite{kadanoff2000statics}, a physical system is coarse grained
by introducing new ``block'' variables which describe some ``effective'' behavior of a block of spins. For example, in the figure,  four adjacent spins are grouped into 
2 x 2 blocks. The system is then described in terms of these new block variables. This scheme is then iterated to create even new block variables
that average over an even larger set of the original spins. Notice the lattice spacing doubles after each iteration. }
\label{Fig:BlockRG}
\end{figure*}

\section{Overview of Variational RG}

In statistical physics, one often considers an ensemble of $N$ binary spins $\{v_i \}$ that can take the values $\pm 1$. The index $i$ labels the position of spin $v_i$ in some lattice. In thermal equilibrium, the probability of a spin configuration is given by the Boltzmann distribution
\be
P(\{v_i \}) =\frac {e^{-{\mathbf H}(\{ v_i\})}}{ Z},
\ee 
where we have defined the Hamiltonian ${\mathbf  H}(\{ v_i\})$, and the partition function 
\be
Z= {\mathrm Tr}_{v_i} e^{-{\mathbf  H}(\{ v_i\})} \equiv  \sum_{v_1, \ldots v_N=\pm 1}e^{-{\mathbf  H}(\{ v_i\})}.
\ee
 Note throughout the paper we set the temperature equal to one, without loss of generality. Typically, the Hamiltonian depends on a set of couplings or parameters, ${\mathbf K} = \{ K_s \}$, that parameterizes the set of all possible Hamiltonians. For example, with binary spins, the  ${\mathbf K}$ could be the couplings describing the spin interactions of various orders:
\be
{\mathbf H}[\{v_i \}] =  -\sum_{i} K_i v_i   - \sum_{ij}  K_{ij} v_i v_j  - \sum_{ijk} K_{ijk} v_i v_j v_k + \ldots.
\label{HamVis}
\ee
Finally, we can define the free energy of the spin system in the standard way:
 \be
 F^{v}= -\log{Z} = -\log{\left({\mathrm Tr}_{v_i} e^{-{\mathbf  H}(\{ v_i\} )}\right)}.
 \ee

The idea behind RG is to find a new coarse-grained description of the spin system where one has ``integrated out'' short distance fluctuations. To this end, let us introduce  $M<N $ new binary spins, $\{h_j \}$. Each of these spins $h_j$ will serve as a coarse-grained degree of freedom where fluctuations on small scales have been averaged out. Typically, such a coarse-graining procedure increases some characteristic length scale describing the system such as the lattice spacing. For example, in the block spin renormalization picture introduced by Kadanoff, each $h_i$ represents the state of a local block of physical spins, $v_i$. Figure \ref{Fig:BlockRG} shows such a block-spin procedure for a two-dimensional spin system on a square lattice, where each $h_i$ represents a $2\times2$ block of visible spins. The result of such a coarse-graining procedure is that the lattice spacing is doubled at each step of the renormalization procedure.

In general, the interactions (statistical correlations) between the $\{ v_i\}$ induce interactions (statistical correlations) between the coarse-grained spins, $\{ h_j\}$. In particular, the coarse-grained system can be described by a new coarse-grained Hamiltonian of the form
\be
{\mathbf{H}^{RG}}[\{h_j\}] =  -\sum_{i}  \tilde{K}_i h_i   - \sum_{ij}  \tilde{K}_{ij} h_i h_j  - \sum_{ijk}  \tilde{K}_{ijk} h_i h_j h_k + \ldots,
\ee
where $\{ \tilde{K} \}$ describe interactions between the hidden spins, $\{ h_j \}$. In the physics literature, such a renormalization transformation is often represented as mapping between couplings, $\{ K\} \to \{ \tilde{K} \}$. Of course, the exact mapping depends on the details of the RG scheme used.

In the variational RG scheme proposed by Kadanoff, the coarse graining procedure is implemented by constructing a function, ${\mathbf T}_\lambda(\{v_i \},\{h_j\})$, that depends on a set of variational parameters $\{\lambda\}$  and encodes (typically pairwise) interactions between the physical and coarse-grained degrees of freedom. After coupling the auxiliary spins $\{h_j\}$ to the physical spins $\{ v_i \}$, one can then integrate out (marginalize over) the visible spins to arrive at a coarse-grained description of the physical system entirely in terms of the $\{ h_j\}$. The function ${\mathbf T}_\lambda(\{v_i \},\{h_j\})$ then naturally defines a Hamiltonian for the $\{ h_j\}$ through the expression
\be
e^{- \mathbf{H}^{RG}_\lambda[\{h_j\}] } \equiv {\mathrm Tr}_{v_i} e^{{\mathbf T}_\lambda(\{v_i\}, \{h_j\})-{\mathbf  H}(\{ v_i\})}.
\label{defhiddenHam}
\ee 
We can also define a free energy for the coarse grained system in the usual way
\be
 F_\lambda^h=  -\log{\left({\mathrm Tr}_{h_i} e^{-{{\mathbf  H}^{RG}}_\lambda(\{ h_i\} )}\right)}.
 \ee

Thus far we  have ignored the problem of choosing  the variational parameters $\lambda$ that define our RG transformation $T_\lambda(\{v_i \},\{h_j\})$. Intuitively, it is clear we should choose $\lambda$ to ensure that the long-distance physical observables of the system are invariant to this coarse graining procedure. This  is done by choosing the parameters $\lambda$ to minimize the free energy difference,  $\Delta F = F_\lambda^h-F^v$, between the physical and coarse grained systems. Notice that
\be
\Delta F=0 \iff  {\mathrm Tr}_{h_j} e^{{\mathbf T}_\lambda(\{v_i\}, \{h_j\})}=1
\ee
Thus, for any {\it exact} RG transformation, we know that 
\be
{\mathrm Tr}_{h_j} e^{{\mathbf T}_\lambda(\{v_i\}, \{h_j\})}=1
\label{exactRG}
\ee

In general, it is not possible to choose the parameters $\lambda$ to satisfy the condition above and various variational schemes (e.g. bond moving) have been proposed to choose $\lambda$  to minimize this $\Delta F$.

\section{RBMs and Deep Neural Networks}

We will show below that this variational RG procedure has a natural interpretation as a deep learning scheme based on a powerful class of energy-based models called Restricted Boltzmann Machines (RBMs) \cite{hinton2006reducing, salakhutdinov2007restricted, larochelle2008classification, smolensky1986information, teh2001rate}.  We will restrict our discussion to RBMs acting on binary data \cite{hinton2006reducing} drawn from some probability distribution, $P(\{v_i\})$, with $\{v_i\}$ binary spins labeled by an index $i=1 \ldots N$.  For example, for black and white images each spin $v_i$ encodes whether a given pixel is on or off and the distribution $P(\{v_i\})$ encodes the statistical properties of the ensemble of images (e.g the set of all handwritten digits in the MNIST dataset).

To model the data distribution, RBMs introduce new hidden spin variables, $\{h_j\}$ ($j=1\ldots M$) that couple to the visible units. The interactions between visible and hidden units are modeled using an energy function of the form
\be
{\mathbf E}(\{v_i\}, \{h_j\}) = \sum_i b_j h_j + \sum_{ij} v_i w_{ij} h_j + \sum_{i} c_i v_i,
\ee
where $\lambda=\{b_j, w_{ij}, c_i\}$ are variational parameters of the model. In terms of this energy function, the joint probability of observing a configuration of hidden and visible spins can be written as
\be
p_{\lambda}(\{v_i\}, \{h_j\})= {e^{-{\mathbf E}(\{v_i\}, \{h_j\})} \over \mathcal{Z}}. 
\label{jointprob}
\ee
This joint distribution also defines a variational distribution for the visible spins
\be
p_{\lambda} (\{v_i\})= \sum_{\{h_j\}} p_{\lambda}(\{v_i\}, \{h_j\}) = {\mathrm Tr}_{h_j} p_{\lambda}(\{v_i\}, \{h_j\})
\ee
as well as a marginal distribution for hidden spins themselves:
\be
p_{\lambda} (\{h_j\})= \sum_{\{v_j\}} p_{\lambda}(\{v_i\}, \{h_j\})={\mathrm Tr}_{v_i} p_{\lambda}(\{v_i\}, \{h_j\}).
\label{marginalhidden}
\ee
Finally, for future reference it will be helpful to define a  ``variational'' RBM Hamiltonian for the visible units:
\be
{p_\lambda(\{ v_i\})} \equiv \frac{ e^{-\mathbf{H}^{RBM}_\lambda[\{v_i\}]}}{\mathcal Z},
 \label{variationalHam}
 \ee
and an RBM Hamiltonian for the hidden units:
\be
p_{\lambda} (\{h_j\}) \equiv  \frac{ e^{-\mathbf{H}^{RBM}_\lambda[\{h_j\}]}}{\mathcal Z}.
\label{RBMhiddenHam}
\ee

Since the objective of the RBM for our purposes is unsupervised learning, the parameters in the RBM are chosen to minimize the Kullback-Leibler divergence  between the true distribution of the data $P(\{ v_i \})$ and the variational distribution $p_{\lambda} (\{v_i\})$:
\be
D_{KL}(P(\{v_i\})|| p_{\lambda}(\{v_i\})= \sum_{\{v_i\}} P(\{v_i\}) \log{\left({P(\{v_i\}) \over p_{\lambda}(\{v_i\})}\right)}.
\ee
Furthermore, notice that when the RBM exactly reproduces the visible data distribution 
\be
D_{KL}(P(\{v_i\})|| p_{\lambda}(\{v_i\}))=0.
\ee
In general it not possible to explicitly minimize the $D_{KL}(P(\{v_i\})|| p_{\lambda}(\{v_i\}))$ and this minimization is usually performed using approximate numerical methods such as contrastive divergence \cite{hinton2002training}. Note that if the number of hidden units is restricted (i.e. less than $2^{N}$), the RBM cannot be made to match an arbitrary distribution exactly \cite{le2008representational}.

In a DNN, RBMs are stacked on top of each other so that, once trained, the hidden layer of one RBM serves as the visible layer of the next RBM. In particular, one can map a configuration of visible spins to a configuration in the hidden layer via the conditional probability distribution, $p_\lambda(\{ h_j\} | \{v_i\})$.  Thus, after training an RBM, we can treat the activities of the hidden layer in response to each visible data sample as data for learning a second layer of hidden spins, and so on.

\section{Mapping Variational RG to Deep Learning}

\begin{figure*}
\center
\includegraphics[width=20cm]{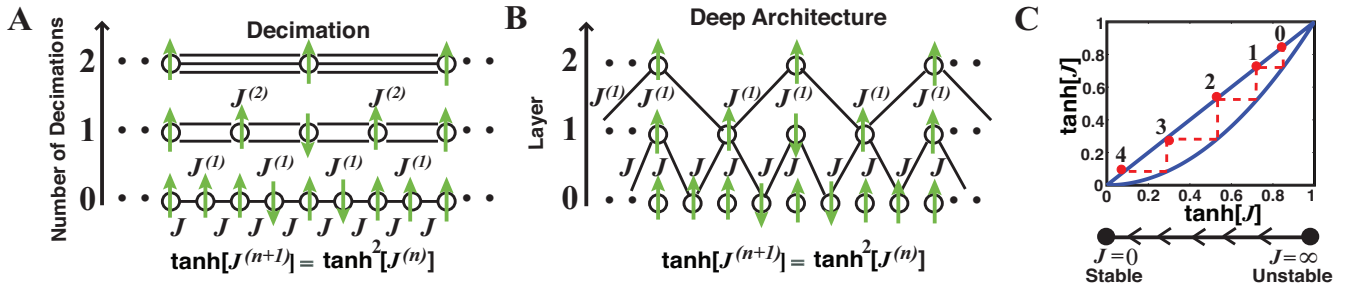}
\caption{{\bf RG and deep learning in the one-dimensional Ising Model.} (A)  A decimation based renormalization transformation
for the ferromagnetic 1-D Ising model. At each step, half the spins are decimated, doubling the effective lattice spacing. After, $n$ successive decimations, the spins can be described using a new 1-D Ising models with a coupling $J^{n}$ between spins. Couplings at a given layer are related to couplings at a previous layer through the square of the hyberbolic tangent function. (B) Decimation-based renormalization transformations can also be realized using the deep architecture where the weights between the $n+1$ and $n$-th hidden layer are given by $J^{n}$.  (C) Visualizing the renormalization group flow of the couplings for 1-D Ferromagnetic Ising model. Under four successive decimations or equivalently as we move up four layers in the deep architecture, the couplings (marked by red dots) get smaller. Eventually, the couplings are attracted to stable fixed point $J=0$.}
\label{Fig:1DIsing}
\end{figure*}

In variational RG,  the couplings between the  hidden and visible spins are encoded by the operators  ${\mathbf T}_\lambda(\{v_i\}, \{h_j\})$. In RBMs, an analogous role is played by the joint energy function ${\mathbf E}(\{v_i\}, \{h_j\})$. In fact, as we will show below, these objects are related through the equation,
\be
\mathbf{T}(\{v_i \},\{h_j \})= -{\mathbf E}(\{v_i\}, \{h_j\})+{\mathbf H}[\{v_i \}],
\label{mapping}
\ee
where ${\mathbf H}[\{v_i \}]$ is the Hamiltonian defined  in Eq. \ref{HamVis} that encodes the data probability distribution $P(\{v_i\})$. This equation defines a one-to-one mapping between the variational RG scheme and RBM based DNNs. 

Using this definition, it is easy to show that  the Hamiltonian $  \mathbf{H}^{RG}_\lambda[\{h_j\}]$, originally  defined in Eq. \ref{defhiddenHam} as the Hamiltonian of the coarse-grained degrees of freedom after performing RG, also describes the hidden spins in the RBM. This is equivalent to the statement that the marginal distribution $p_\lambda(\{h_j\})$  describing the hidden spins of the RBM is of the Boltzmann form with a Hamiltonian  $ \mathbf{H}^{RG}_\lambda[\{h_j\}]$.  To prove this,  we divide both sides of  Eq. \ref{defhiddenHam} by ${\mathcal Z}$ to get
 \be
\frac{e^{- \mathbf{H}^{RG}_\lambda[\{h_j\}] } }{\mathcal Z} = \frac{ {\mathrm Tr}_{v_i} e^{{\mathbf T}_\lambda(\{v_i\}, \{h_j\})-{\mathbf  H}(\{ v_i\})}}{\mathcal Z}.
\ee
Substituting Eq. \ref{mapping} into this equation yields
\be
\frac{e^{- \mathbf{H}^{RG}_\lambda[\{h_j\}] }}{\mathcal Z} = {\mathrm Tr}_{v_i} {e^{-{\mathbf E}(\{v_i\}, \{h_j\})} \over \mathcal{Z}}  = p_\lambda(\{h_j\}).
\ee
Substituting Eq. \ref{RBMhiddenHam} into the right-hand side  yields the desired result
\be
\mathbf{H}^{RG}_\lambda[\{h_j\}]=\mathbf{H}^{RBM}_\lambda[\{h_j\}].
\ee

These results also provide a natural interpretation for variational RG entirely in the language of probability theory. The operator ${\mathbf T}_\lambda(\{v_i\}, \{h_j\})$ can be viewed as a variational approximation for the conditional probability of the hidden spins given the visible spins. To see this, notice that  
\bea
e^{ \mathbf{T}(\{v_i \},\{h_j \})}&=& e^{-{\mathbf E}(\{v_i\}, \{h_j\})+{\mathbf H}[\{v_i \}] } \nonumber \\
&=& \frac{p_{\lambda}(\{v_i\}, \{h_j\})}{p_\lambda(\{v_i\})} e^{{\mathbf H}[\{v_i \}] - \mathbf{H}^{RBM}_\lambda[\{v_i\}] } \nonumber \\
&=& p_\lambda( \{h_j\} | \{v_i\}) e^{{\mathbf H}[\{v_i \}] -\mathbf{H}^{RBM}_\lambda[\{v_i\}] }
\eea
where in going from the first the line to the second line we have used Eqs.  \ref{jointprob} and  \ref{variationalHam}.
This implies that when an RG can be performed exactly (i.e. the  RG transformation satisfies the equality $ {\mathrm Tr}_{h_j} e^{{\mathbf T}_\lambda(\{v_i\}, \{h_j\})} =1$), the variational Hamiltonian is identical to the true  Hamiltonian describing the data, ${\mathbf H}[\{v_i \}] = \mathbf{H}^{RBM}_\lambda[\{v_i\}]$ and $ \mathbf{T}(\{v_i \},\{h_j \})$ is exactly the conditional probability.  In the language of probability theory, this means that the variational distribution $p_\lambda(\{v_i\})$ exactly reproduces the true data distribution $P(\{v_i \})$ and $D_{KL}(P(\{v_i\})|| p_{\lambda}(\{v_i\})=0$.  

In general, it is not possible to perform the variational RG transformation exactly. Instead, one constructs a family of variational approximations for the exact RG transform  \cite{kadanoff1976variational,kadanoff2000statics, efrati2014real}. The discussion above makes it clear that these variational distributions work at the level of the Hamiltonians and Free Energies. In contrast, in the Machine Learning literature, these variational approximations are usually made by minimizing the KL divergence $D_{KL}(P(\{v_i\})|| p_{\lambda}(\{v_i\})=0$. Thus, the two approaches employ  distinct variational approximation schemes for coarse graining. Finally, notice that the correspondence does not rely on the explicit form of the energy $E(\{h_j\}, \{v_j\})$ and hence holds for any Boltzmann Machine. 

\section{Examples}

To gain intuition about the mapping between RG and deep learning, it is helpful to consider some simple examples in detail. We begin by examining the one-dimensional nearest-neighbor Ising model where the RG transformation can be carried out exactly. We then numerically explore the two-dimensional nearest-neighbor Ising model using an RBM-based deep learning architecture.

\subsection{One dimensional Ising Model}

The one-dimensional Ising model describes a collection of binary spins $\{ v_i \}$ organized along a one-dimensional lattice with lattice spacing $a$. Such a system is described by a Hamiltonian of the form
\be
H=-J \sum_{i} v_i v_{i+1},
\ee
where $J$ is a ferromagnetic coupling that energetically favors configurations where neighboring spins align. To perform a RG transformation, we decimate (marginalize over) every other spin. This doubles the lattice spacing $a \rightarrow 2a$ and results in a new effective interaction $J^{(1)}$ between spins (see Figure \ref{Fig:1DIsing}). If we denote the coupling after performing $n$ successive RG transformations by $J^{(n)}$, then a standard calculation shows that these coefficients satisfy the RG equations 
\be
\tanh{[J^{(n+1)}]} = \tanh^2{[J^{(n)}]},
\ee
where we have defined $J^{(0)}=J$ \cite{kadanoff2000statics}. This recursion relationship can be visualized as a one-dimensional flow in the coupling space $J$ from $J=\infty$ to $J=0$. Thus, after performing RG the interactions become weaker and weaker and $J \rightarrow 0$ as $n \rightarrow \infty$.

This RG transformation also naturally gives rise to the deep learning architecture shown in Figure \ref{Fig:1DIsing}. The spins at a given layer of the DNN have a natural interpretation as the decimated spins when performing the RG transformation in the layer below.  Notice that the coupled spins in the bottom two layers of the DNNs in Fig. \ref{Fig:1DIsing}B form an ``effective" one-dimensional chain isomorphic to the original spin chain. Thus, marginalizing over spins in the bottom layer in the DNN is identical to decimating every other spin in the original spin systems. This implies that the ``hidden'' spins in the second layer of the DNN are also described by the RG transformed Hamiltonian with a coupling  $J^{(1)}$ between neighboring spins. Repeating this argument for  spins coupled between the second and third layers and so on, one obtains the deep learning architecture shown in  Fig. \ref{Fig:1DIsing}B which implements decimation. 

The advantage of the simple deep architecture presented here is that it is easy to interpret and requires no calculations to construct. However, an important shortcoming is that it contains no information about half of the visible spins, namely the spins that do not couple to the hidden layer.

\subsection{Two dimensional Ising Model}

\begin{figure*}
\center
\includegraphics[width=15cm]{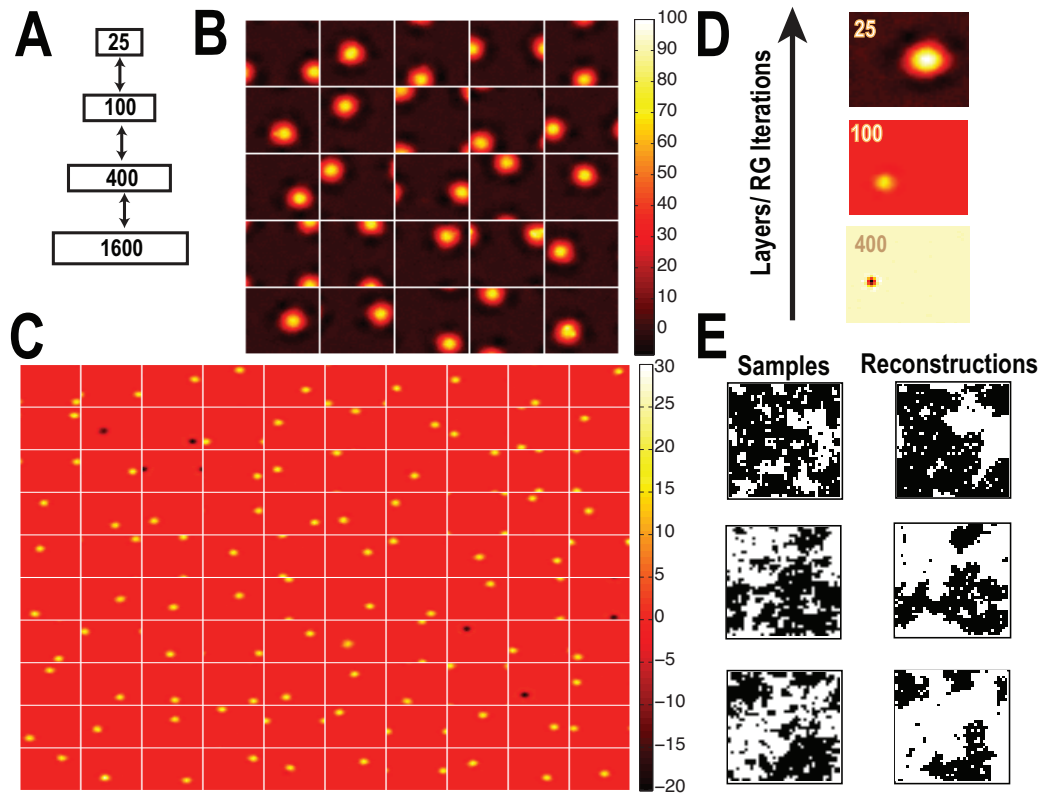}
\caption{{\bf Deep learning the 2D Ising model}  {\bf A} Deep Neural Network with four layers of size 1600, 400, 100, and 25 spins was trained using samples drawn from a 2D Ising model slightly above the critical temperature, $J/(k_BT)=0.408$. {\bf B} Visualization of the effective receptive fields for the top layer of spins. Each 40 by 40 pixel image depicts the effective receptive field  of one of the 25 spins in the top layer (see material and methods){\bf C} Visualization of effective receptive fields for each of the 100 spins in the middle layer calculated as in B. {\bf D} The effective receptive fields get larger as one moves up the Deep Neural Network. This is consistent with what is expected from the successive application of block renormalization. {\bf E} Three representative samples drawn from the 2D Ising model at $J=0.408$ and their reconstruction from the trained DNN. Samples were reconstructed from DNNs as in \cite{hinton2006reducing}.}
\label{Fig:2DIsing}
\end{figure*}

We next applied deep learning techniques to numerically coarse-grain the two-dimensional nearest-neighbor Ising model on a square lattice. This model is described by a Hamiltonian of the form
\be
H[\{v_i \}] = -J \sum_{\left<i j\right>}  v_i v_j, 
\ee
where $\left<i j\right>$ indicates that $i$ and $j$ are nearest neighbors and $J$ is a ferromagnetic coupling that favors configurations where neighboring spins align. Unlike the one-dimensional Ising model, the two dimensional Ising model has a phase transition that occurs when $J/(k_BT)= 0.4352$ (recall we have set $\beta=T^{-1}=1$). At the phase transition, the characteristic length scale of the system, the correlation length, diverges. For this reason, near a critical point the system can be productively coarse-grained using a procedure similar to Kadanoff's block spin renormalization (see Fig. \ref{Fig:BlockRG}) \cite{kadanoff2000statics}.

Inspired by our mapping between variational RG and DNNs, we applied standard deep learning techniques to samples generated from the $2$D Ising model for $J=0.408$, just above the critical temperature. $20,000$ samples were generated from a periodic $40\times40$ $2$D Ising model using standard equilibrium Monte Carlo techniques and served as input to an RBM-based deep neural network of four layers with $1600$, $400$, $100$, and $25$ spins respectively (see Fig. \ref{Fig:2DIsing}A). We furthermore imposed an L$1$ penalty on the weights between layers in the RBM and trained the network using contrastive divergence  \cite{hinton2002training} (see Materials and Methods).  The L$1$ penalty serves as a sparsity promoting regularizer that encourages weights in the RBM to be zero and prevents overfitting due to the finite number of samples. In practice, it ensures that visible and hidden spins interact with only a small subset of all the spins in an RBM. (Note that we did not use a convolutional network that explicitly builds in spatial locality or translational invariance.)

The architecture of the resulting DNN suggests that it is implementing a coarse-graining scheme similar to block spin renormalization (see Fig. \ref{Fig:2DIsing}). Each spin in a hidden layer couples to a local block of spins in the layer below. This iterative blocking is consistent with Kadanoff's intuitive picture of how coarse-graining should be implemented near the critical point. Moreover, the size of the blocks coupling to each hidden unit in a layer are of approximately the same size (Fig. \ref{Fig:2DIsing}B,C), and the characteristic size is increasing with layer (Fig. \ref{Fig:2DIsing}D). \emph{Surprisingly, this local block spin structure emerges from the training process, suggesting the DNN is self-organizing to implement block spin renormalization.} Furthermore, as shown in Fig. \ref{Fig:2DIsing}E, reconstructions from the coarse grained DNN can qualitatively reproduce the macroscopic features of individual samples despite having only $25$ spins in the top layer, a compression ratio of $64$.

\section{Discussion}

Deep learning is one of the most successful paradigms for  unsupervised learning to emerge over the last ten years. The enormous success of deep learning techniques at a variety of practical machine learning tasks ranging from voice recognition to image classification raises natural questions about its theoretical underpinnings. Here, we have demonstrated that there is a one-to-one mapping between RBM-based Deep Neural Networks and the variational renormalization group. We illustrated this mapping by analytically constructing a DNN for the 1D Ising model and numerically examining the 2D Ising model. Surprisingly, we found that these DNNs self organize to implement a coarse-graining procedure reminiscent of Kadanoff block renormalization. This suggests that deep learning may be implementing a generalized RG-like scheme to learn important features from data. 

RG plays a central role in our modern understanding of statistical physics and quantum field theory. A central finding of RG is that the long distance physics of many disparate physical systems are dominated by the same long distance fixed points. This gives rise to the idea of universality -- many microscopically dissimilar systems exhibit macroscopically similar properties at long distances  Physicists have developed elaborate technical machinery for exploiting fixed points and universality to identify the salient long distance features of physics systems. It will be interesting to see, what, if any of this more complex machinery can be imported to deep learning. A potential obstacle for importing ideas from physics into the deep learning framework is that RG is commonly applied to physical systems with many symmetries.  This is in contrast to deep learning which is often applied to data with limited structure. 
 
Recently, it was suggested that modern RG techniques developed in the context of quantum systems such as matrix product states and tensor networks have a natural interpretation in terms of variational RG \cite{ efrati2014real}. These new techniques exploit ideas such as entanglement entropy and disentanglers which create a features with a minimum amount of redundancy. It is an open question to see whether these ideas can be imported into deep learning algorithms. Our mapping also suggests a route for applying real space renormalization techniques to complicated physical systems. Real space renormalization techniques such as variational RG have often been limited by their inability to make good approximations. Techniques from deep learning may represent a possible route for overcoming these problems.

\begin{appendix}
\section{Learning Deep Architecture for the Two-dimensional Ising Model} 
Details are given in the {\it SI Materials and Methods}. Stacked RBMs were trained with a variant of the code from \cite{hinton2006reducing}. This code is available at https://code.google.com/p/matrbm/.  In particular, only the unsupervised learning phase was performed. Individual RBMs were trained with contrastive divergence for $200$ epochs, with momentum $0.5$ using mini-batches of size $100$ on $40,000$ total samples from the $2D$ Ising model with $J=0.408$. Additionally, $L1$ regularization was implemented, with strength $2\times10^{-4}$, instead of weight decay. This L1 regularization strength was chosen to ensure that one could not have all-to-all couplings between layers in the DNN.
Reconstructions were performed as in \cite{hinton2006reducing}. See Supplementary files for a Matlab variable containing the learned model.

\section{Visualizing Effective Receptive Fields}

The effective receptive field is a way to visualize which spins in the visible layer that coupled to a given spin in one of the  hidden layers. We denote the effective receptive field matrix of layer $l$ by $r^{(l)}$ and the number of spins in layer $l$ by $n^{(l)}$, with the visible layer corresponding to $l=0$. Each column in $r^{(l)}$ is a vector that encodes the receptive field of a single spin in hidden layer $l$. It can be computed by convoluting the weight matrices $W^{(l)}$ encoding the weights $w_{ij}$ between the spins in layers $l-1$ and $l$ . To compute $r^{(l)}$ first we set $r^{(1)}=W^{(1)}$ and used  the recursion relationship $r^l=r^{(l-1)}W^{(l)}$ for $l>1$. Thus, the effective receptive field of a spin is a measure of how much that hidden spin influences the spins in the visible layer. 
\end{appendix}

\begin{acknowledgments}
PM is grateful to Charles K. Fisher for useful conversations. We are also grateful to Javad Noorbakhsh and Alex Lang for comments on the manuscript. This work was partially supported by Simons Foundation Investigator Award in the Mathematical Modeling of Living Systems and a Sloan Research Fellowship (to P.M). DJS was partially supported by NIH Grant K25 GM098875.
\end{acknowledgments}

\bibliography{refsmain}

\begin{thebibliography}{24}%
\makeatletter
\providecommand \@ifxundefined [1]{%
 \@ifx{#1\undefined}
}%
\providecommand \@ifnum [1]{%
 \ifnum #1\expandafter \@firstoftwo
 \else \expandafter \@secondoftwo
 \fi
}%
\providecommand \@ifx [1]{%
 \ifx #1\expandafter \@firstoftwo
 \else \expandafter \@secondoftwo
 \fi
}%
\providecommand \natexlab [1]{#1}%
\providecommand \enquote  [1]{``#1''}%
\providecommand \bibnamefont  [1]{#1}%
\providecommand \bibfnamefont [1]{#1}%
\providecommand \citenamefont [1]{#1}%
\providecommand \href@noop [0]{\@secondoftwo}%
\providecommand \href [0]{\begingroup \@sanitize@url \@href}%
\providecommand \@href[1]{\@@startlink{#1}\@@href}%
\providecommand \@@href[1]{\endgroup#1\@@endlink}%
\providecommand \@sanitize@url [0]{\catcode `\\12\catcode `\$12\catcode
  `\&12\catcode `\#12\catcode `\^12\catcode `\_12\catcode `\%12\relax}%
\providecommand \@@startlink[1]{}%
\providecommand \@@endlink[0]{}%
\providecommand \url  [0]{\begingroup\@sanitize@url \@url }%
\providecommand \@url [1]{\endgroup\@href {#1}{\urlprefix }}%
\providecommand \urlprefix  [0]{URL }%
\providecommand \Eprint [0]{\href }%
\@ifxundefined \urlstyle {%
  \providecommand \doi  [0]{\begingroup \@sanitize@url \@doi}%
  \providecommand \@doi [1]{\endgroup \@@startlink {\doibase
  #1}doi:\discretionary {}{}{}#1\@@endlink }%
}{%
  \providecommand \doi  [0]{doi:\discretionary{}{}{}\begingroup
  \urlstyle{rm}\Url }%
}%
\providecommand \doibase [0]{http://dx.doi.org/}%
\providecommand \Doi [0]{\begingroup \@sanitize@url \@Doi }%
\providecommand \@Doi  [1]{\endgroup\@@startlink{\doibase#1}\@@Doi}%
\providecommand \@@Doi [1]{#1\@@endlink}%
\providecommand \selectlanguage [0]{\@gobble}%
\providecommand \bibinfo  [0]{\@secondoftwo}%
\providecommand \bibfield  [0]{\@secondoftwo}%
\providecommand \translation [1]{[#1]}%
\providecommand \BibitemOpen [0]{}%
\providecommand \bibitemStop [0]{}%
\providecommand \bibitemNoStop [0]{.\EOS\space}%
\providecommand \EOS [0]{\spacefactor3000\relax}%
\providecommand \BibitemShut  [1]{\csname bibitem#1\endcsname}%
\bibitem [{\citenamefont {Bengio}(2009)}]{bengio2009learning}%
  \BibitemOpen
  \bibfield  {author} {\bibinfo {author} {\bibfnamefont {Y.}~\bibnamefont
  {Bengio}},\ }\href@noop {} {\bibfield  {journal} {\bibinfo  {journal}
  {Foundations and trends{\textregistered} in Machine Learning},\ }\textbf
  {\bibinfo {volume} {2}},\ \bibinfo {pages} {1} (\bibinfo {year}
  {2009})}\BibitemShut {NoStop}%
\bibitem [{\citenamefont {Le}\ \emph {et~al.}(2012)\citenamefont {Le},
  \citenamefont {Ranzato}, \citenamefont {Monga}, \citenamefont {Devin},
  \citenamefont {Chen}, \citenamefont {Corrado}, \citenamefont {Dean},\ and\
  \citenamefont {Ng}}]{38115}%
  \BibitemOpen
  \bibfield  {author} {\bibinfo {author} {\bibfnamefont {Q.}~\bibnamefont
  {Le}}, \bibinfo {author} {\bibfnamefont {M.}~\bibnamefont {Ranzato}},
  \bibinfo {author} {\bibfnamefont {R.}~\bibnamefont {Monga}}, \bibinfo
  {author} {\bibfnamefont {M.}~\bibnamefont {Devin}}, \bibinfo {author}
  {\bibfnamefont {K.}~\bibnamefont {Chen}}, \bibinfo {author} {\bibfnamefont
  {G.}~\bibnamefont {Corrado}}, \bibinfo {author} {\bibfnamefont
  {J.}~\bibnamefont {Dean}}, \ and\ \bibinfo {author} {\bibfnamefont
  {A.}~\bibnamefont {Ng}},\ }\href@noop {} {\bibfield  {journal} {\bibinfo
  {journal} {International Conference in Machine Learning}} (\bibinfo {year}
  {2012})}\BibitemShut {NoStop}%
\bibitem [{\citenamefont {Krizhevsky}\ \emph {et~al.}(2012)\citenamefont
  {Krizhevsky}, \citenamefont {Sutskever},\ and\ \citenamefont
  {Hinton}}]{imagenet}%
  \BibitemOpen
  \bibfield  {author} {\bibinfo {author} {\bibfnamefont {A.}~\bibnamefont
  {Krizhevsky}}, \bibinfo {author} {\bibfnamefont {I.}~\bibnamefont
  {Sutskever}}, \ and\ \bibinfo {author} {\bibfnamefont {G.~E.}\ \bibnamefont
  {Hinton}},\ }\href@noop {} {\bibfield  {journal} {\bibinfo  {journal}
  {Advances in Neural Information Processing Systems 25},\ \bibinfo {pages}
  {1097}} (\bibinfo {year} {2012})}\BibitemShut {NoStop}%
\bibitem [{\citenamefont {Hinton}\ \emph {et~al.}(2012)\citenamefont {Hinton},
  \citenamefont {Deng}, \citenamefont {Yu}, \citenamefont {Dahl}, \citenamefont
  {Mohamed}, \citenamefont {Jaitly}, \citenamefont {Senior}, \citenamefont
  {Vanhoucke}, \citenamefont {Nguyen}, \citenamefont {Sainath},\ and\
  \citenamefont {Kingsbury}}]{speechrecognition}%
  \BibitemOpen
  \bibfield  {author} {\bibinfo {author} {\bibfnamefont {G.}~\bibnamefont
  {Hinton}}, \bibinfo {author} {\bibfnamefont {L.}~\bibnamefont {Deng}},
  \bibinfo {author} {\bibfnamefont {D.}~\bibnamefont {Yu}}, \bibinfo {author}
  {\bibfnamefont {G.}~\bibnamefont {Dahl}}, \bibinfo {author} {\bibfnamefont
  {A.}~\bibnamefont {Mohamed}}, \bibinfo {author} {\bibfnamefont
  {N.}~\bibnamefont {Jaitly}}, \bibinfo {author} {\bibfnamefont
  {A.}~\bibnamefont {Senior}}, \bibinfo {author} {\bibfnamefont
  {V.}~\bibnamefont {Vanhoucke}}, \bibinfo {author} {\bibfnamefont
  {P.}~\bibnamefont {Nguyen}}, \bibinfo {author} {\bibfnamefont
  {T.}~\bibnamefont {Sainath}}, \ and\ \bibinfo {author} {\bibfnamefont
  {B.}~\bibnamefont {Kingsbury}},\ }\href@noop {} {\bibfield  {journal}
  {\bibinfo  {journal} {Signal Processing Magazine},\ }\textbf {\bibinfo
  {volume} {29}},\ \bibinfo {pages} {82} (\bibinfo {year} {2012})}\BibitemShut
  {NoStop}%
\bibitem [{\citenamefont {Sarikaya}\ \emph {et~al.}(2014)\citenamefont
  {Sarikaya}, \citenamefont {Hinton},\ and\ \citenamefont {Deoras}}]{deepnlp}%
  \BibitemOpen
  \bibfield  {author} {\bibinfo {author} {\bibfnamefont {R.}~\bibnamefont
  {Sarikaya}}, \bibinfo {author} {\bibfnamefont {G.}~\bibnamefont {Hinton}}, \
  and\ \bibinfo {author} {\bibfnamefont {A.}~\bibnamefont {Deoras}},\
  }\href@noop {} {\bibfield  {journal} {\bibinfo  {journal} {IEEE Transactions
  on Audio Speech and Language Processing}} (\bibinfo {year}
  {2014})}\BibitemShut {NoStop}%
\bibitem [{\citenamefont {Hinton}\ and\ \citenamefont
  {Salakhutdinov}(2006)}]{hinton2006reducing}%
  \BibitemOpen
  \bibfield  {author} {\bibinfo {author} {\bibfnamefont {G.~E.}\ \bibnamefont
  {Hinton}}\ and\ \bibinfo {author} {\bibfnamefont {R.~R.}\ \bibnamefont
  {Salakhutdinov}},\ }\href@noop {} {\bibfield  {journal} {\bibinfo  {journal}
  {Science},\ }\textbf {\bibinfo {volume} {313}},\ \bibinfo {pages} {504}
  (\bibinfo {year} {2006})}\BibitemShut {NoStop}%
\bibitem [{\citenamefont {Bengio}\ and\ \citenamefont
  {Yann}(2007)}]{bengiolecun}%
  \BibitemOpen
  \bibfield  {author} {\bibinfo {author} {\bibfnamefont {Y.}~\bibnamefont
  {Bengio}}\ and\ \bibinfo {author} {\bibfnamefont {L.}~\bibnamefont {Yann}},\
  }\href@noop {} {\bibfield  {journal} {\bibinfo  {journal} {Large-scale kernel
  machines},\ }\textbf {\bibinfo {volume} {34}},\ \bibinfo {pages} {1}
  (\bibinfo {year} {2007})}\BibitemShut {NoStop}%
\bibitem [{\citenamefont {Le~Roux}\ and\ \citenamefont
  {Bengio}(2010)}]{le2010}%
  \BibitemOpen
  \bibfield  {author} {\bibinfo {author} {\bibfnamefont {N.}~\bibnamefont
  {Le~Roux}}\ and\ \bibinfo {author} {\bibfnamefont {Y.}~\bibnamefont
  {Bengio}},\ }\href@noop {} {\bibfield  {journal} {\bibinfo  {journal} {Neural
  Computation},\ }\textbf {\bibinfo {volume} {22}},\ \bibinfo {pages} {2192}
  (\bibinfo {year} {2010})}\BibitemShut {NoStop}%
\bibitem [{\citenamefont {Le~Roux}\ and\ \citenamefont
  {Bengio}(2008)}]{le2008representational}%
  \BibitemOpen
  \bibfield  {author} {\bibinfo {author} {\bibfnamefont {N.}~\bibnamefont
  {Le~Roux}}\ and\ \bibinfo {author} {\bibfnamefont {Y.}~\bibnamefont
  {Bengio}},\ }\href@noop {} {\bibfield  {journal} {\bibinfo  {journal} {Neural
  Computation},\ }\textbf {\bibinfo {volume} {20}},\ \bibinfo {pages} {1631}
  (\bibinfo {year} {2008})}\BibitemShut {NoStop}%
\bibitem [{\citenamefont {Bengio}\ \emph {et~al.}(2007)\citenamefont {Bengio},
  \citenamefont {Lamblin}, \citenamefont {Popovici},\ and\ \citenamefont
  {Larochelle}}]{bengio2007greedy}%
  \BibitemOpen
  \bibfield  {author} {\bibinfo {author} {\bibfnamefont {Y.}~\bibnamefont
  {Bengio}}, \bibinfo {author} {\bibfnamefont {P.}~\bibnamefont {Lamblin}},
  \bibinfo {author} {\bibfnamefont {D.}~\bibnamefont {Popovici}}, \ and\
  \bibinfo {author} {\bibfnamefont {H.}~\bibnamefont {Larochelle}},\
  }\href@noop {} {\bibfield  {journal} {\bibinfo  {journal} {Advances in neural
  information processing systems},\ }\textbf {\bibinfo {volume} {19}},\
  \bibinfo {pages} {153} (\bibinfo {year} {2007})}\BibitemShut {NoStop}%
\bibitem [{\citenamefont {Wilson}\ and\ \citenamefont
  {Kogut}(1974)}]{wilson1974renormalization}%
  \BibitemOpen
  \bibfield  {author} {\bibinfo {author} {\bibfnamefont {K.~G.}\ \bibnamefont
  {Wilson}}\ and\ \bibinfo {author} {\bibfnamefont {J.}~\bibnamefont {Kogut}},\
  }\href@noop {} {\bibfield  {journal} {\bibinfo  {journal} {Physics Reports},\
  }\textbf {\bibinfo {volume} {12}},\ \bibinfo {pages} {75} (\bibinfo {year}
  {1974})}\BibitemShut {NoStop}%
\bibitem [{\citenamefont {Wilson}(1983)}]{wilson1983renormalization}%
  \BibitemOpen
  \bibfield  {author} {\bibinfo {author} {\bibfnamefont {K.~G.}\ \bibnamefont
  {Wilson}},\ }\href@noop {} {\bibfield  {journal} {\bibinfo  {journal}
  {Reviews of Modern Physics},\ }\textbf {\bibinfo {volume} {55}},\ \bibinfo
  {pages} {583} (\bibinfo {year} {1983})}\BibitemShut {NoStop}%
\bibitem [{\citenamefont {Cardy}(1996)}]{cardy1996scaling}%
  \BibitemOpen
  \bibfield  {author} {\bibinfo {author} {\bibfnamefont {J.}~\bibnamefont
  {Cardy}},\ }\href@noop {} {\emph {\bibinfo {title} {Scaling and
  renormalization in statistical physics}}},\ Vol.~\bibinfo {volume} {5}\
  (\bibinfo  {publisher} {Cambridge University Press},\ \bibinfo {year}
  {1996})\BibitemShut {NoStop}%
\bibitem [{\citenamefont {Kadanoff}(2000)}]{kadanoff2000statics}%
  \BibitemOpen
  \bibfield  {author} {\bibinfo {author} {\bibfnamefont {L.~P.}\ \bibnamefont
  {Kadanoff}},\ }\href@noop {} {\emph {\bibinfo {title} {Statics, Dynamics and
  Renormalization}}}\ (\bibinfo  {publisher} {World Scientific},\ \bibinfo
  {year} {2000})\BibitemShut {NoStop}%
\bibitem [{\citenamefont {Goldenfeld}(1992)}]{goldenfeld1992lectures}%
  \BibitemOpen
  \bibfield  {author} {\bibinfo {author} {\bibfnamefont {N.}~\bibnamefont
  {Goldenfeld}},\ }\href@noop {} { (\bibinfo {year} {1992})}\BibitemShut
  {NoStop}%
\bibitem [{\citenamefont {Kadanoff}\ \emph {et~al.}(1976)\citenamefont
  {Kadanoff}, \citenamefont {Houghton},\ and\ \citenamefont
  {Yalabik}}]{kadanoff1976variational}%
  \BibitemOpen
  \bibfield  {author} {\bibinfo {author} {\bibfnamefont {L.~P.}\ \bibnamefont
  {Kadanoff}}, \bibinfo {author} {\bibfnamefont {A.}~\bibnamefont {Houghton}},
  \ and\ \bibinfo {author} {\bibfnamefont {M.~C.}\ \bibnamefont {Yalabik}},\
  }\href@noop {} {\bibfield  {journal} {\bibinfo  {journal} {Journal of
  Statistical Physics},\ }\textbf {\bibinfo {volume} {14}},\ \bibinfo {pages}
  {171} (\bibinfo {year} {1976})}\BibitemShut {NoStop}%
\bibitem [{\citenamefont {Efrati}\ \emph {et~al.}(2014)\citenamefont {Efrati},
  \citenamefont {Wang}, \citenamefont {Kolan},\ and\ \citenamefont
  {Kadanoff}}]{efrati2014real}%
  \BibitemOpen
  \bibfield  {author} {\bibinfo {author} {\bibfnamefont {E.}~\bibnamefont
  {Efrati}}, \bibinfo {author} {\bibfnamefont {Z.}~\bibnamefont {Wang}},
  \bibinfo {author} {\bibfnamefont {A.}~\bibnamefont {Kolan}}, \ and\ \bibinfo
  {author} {\bibfnamefont {L.~P.}\ \bibnamefont {Kadanoff}},\ }\href@noop {}
  {\bibfield  {journal} {\bibinfo  {journal} {Reviews of Modern Physics},\
  }\textbf {\bibinfo {volume} {86}},\ \bibinfo {pages} {647} (\bibinfo {year}
  {2014})}\BibitemShut {NoStop}%
\bibitem [{\citenamefont {Bengio}\ \emph {et~al.}(2013)\citenamefont {Bengio},
  \citenamefont {Courville},\ and\ \citenamefont
  {Vincent}}]{bengio2013representation}%
  \BibitemOpen
  \bibfield  {author} {\bibinfo {author} {\bibfnamefont {Y.}~\bibnamefont
  {Bengio}}, \bibinfo {author} {\bibfnamefont {A.}~\bibnamefont {Courville}}, \
  and\ \bibinfo {author} {\bibfnamefont {P.}~\bibnamefont {Vincent}},\
  }\href@noop {} {\bibfield  {journal} {\bibinfo  {journal} {Pattern Analysis
  and Machine Intelligence, IEEE Transactions on},\ }\textbf {\bibinfo {volume}
  {35}},\ \bibinfo {pages} {1798} (\bibinfo {year} {2013})}\BibitemShut
  {NoStop}%
\bibitem [{\citenamefont {Hinton}\ \emph {et~al.}(2006)\citenamefont {Hinton},
  \citenamefont {Osindero},\ and\ \citenamefont {Teh}}]{hinton2006fast}%
  \BibitemOpen
  \bibfield  {author} {\bibinfo {author} {\bibfnamefont {G.}~\bibnamefont
  {Hinton}}, \bibinfo {author} {\bibfnamefont {S.}~\bibnamefont {Osindero}}, \
  and\ \bibinfo {author} {\bibfnamefont {Y.-W.}\ \bibnamefont {Teh}},\
  }\href@noop {} {\bibfield  {journal} {\bibinfo  {journal} {Neural
  computation},\ }\textbf {\bibinfo {volume} {18}},\ \bibinfo {pages} {1527}
  (\bibinfo {year} {2006})}\BibitemShut {NoStop}%
\bibitem [{\citenamefont {Salakhutdinov}\ \emph {et~al.}(2007)\citenamefont
  {Salakhutdinov}, \citenamefont {Mnih},\ and\ \citenamefont
  {Hinton}}]{salakhutdinov2007restricted}%
  \BibitemOpen
  \bibfield  {author} {\bibinfo {author} {\bibfnamefont {R.}~\bibnamefont
  {Salakhutdinov}}, \bibinfo {author} {\bibfnamefont {A.}~\bibnamefont {Mnih}},
  \ and\ \bibinfo {author} {\bibfnamefont {G.}~\bibnamefont {Hinton}},\ }in\
  \href@noop {} {\emph {\bibinfo {booktitle} {Proceedings of the 24th
  international conference on Machine learning}}}\ (\bibinfo {organization}
  {ACM},\ \bibinfo {year} {2007})\ pp.\ \bibinfo {pages} {791--798}\BibitemShut
  {NoStop}%
\bibitem [{\citenamefont {Larochelle}\ and\ \citenamefont
  {Bengio}(2008)}]{larochelle2008classification}%
  \BibitemOpen
  \bibfield  {author} {\bibinfo {author} {\bibfnamefont {H.}~\bibnamefont
  {Larochelle}}\ and\ \bibinfo {author} {\bibfnamefont {Y.}~\bibnamefont
  {Bengio}},\ }in\ \href@noop {} {\emph {\bibinfo {booktitle} {Proceedings of
  the 25th international conference on Machine learning}}}\ (\bibinfo
  {organization} {ACM},\ \bibinfo {year} {2008})\ pp.\ \bibinfo {pages}
  {536--543}\BibitemShut {NoStop}%
\bibitem [{\citenamefont {Smolensky}(1986)}]{smolensky1986information}%
  \BibitemOpen
  \bibfield  {author} {\bibinfo {author} {\bibfnamefont {P.}~\bibnamefont
  {Smolensky}},\ }\href@noop {} { (\bibinfo {year} {1986})}\BibitemShut
  {NoStop}%
\bibitem [{\citenamefont {Teh}\ and\ \citenamefont
  {Hinton}(2001)}]{teh2001rate}%
  \BibitemOpen
  \bibfield  {author} {\bibinfo {author} {\bibfnamefont {Y.~W.}\ \bibnamefont
  {Teh}}\ and\ \bibinfo {author} {\bibfnamefont {G.~E.}\ \bibnamefont
  {Hinton}},\ }\href@noop {} {\bibfield  {journal} {\bibinfo  {journal}
  {Advances in neural information processing systems},\ \bibinfo {pages} {908}}
  (\bibinfo {year} {2001})}\BibitemShut {NoStop}%
\bibitem [{\citenamefont {Hinton}(2002)}]{hinton2002training}%
  \BibitemOpen
  \bibfield  {author} {\bibinfo {author} {\bibfnamefont {G.~E.}\ \bibnamefont
  {Hinton}},\ }\href@noop {} {\bibfield  {journal} {\bibinfo  {journal} {Neural
  computation},\ }\textbf {\bibinfo {volume} {14}},\ \bibinfo {pages} {1771}
  (\bibinfo {year} {2002})}\BibitemShut {NoStop}%
\end{thebibliography}%

\end{document}